%% file: arxiv_cvpr.tex
\documentclass[10pt,twocolumn,letterpaper]{article}
\usepackage[pagenumbers]{cvpr}

\input{preamble}

\renewcommand{\towa}[1]{}
\newcommand{\changyue}[1]{}
\newcommand{\gc}[1]{}
\newcommand{\dx}[1]{}

\definecolor{cvprblue}{rgb}{0.21,0.49,0.74}
\usepackage[breaklinks,colorlinks,allcolors=cvprblue]{hyperref}
\def\confName{}
\def\confYear{}

\definecolor{prioblue}{HTML}{245A81}
\title{\texorpdfstring{\textcolor{prioblue}{PRIO}}{PRIO}: \textcolor{prioblue}{P}re-trained \textcolor{prioblue}{R}epresentation for \textcolor{prioblue}{I}nteractive \textcolor{prioblue}{O}bjects in 4D Worlds}

\author{
\begin{tabular}{c@{\hspace{1.6em}}c@{\hspace{1.6em}}c@{\hspace{1.6em}}c}
Shixun Huang$^{1}$ & Lifu Wei$^{1}$ & Yue Chang$^{2,*}$ & Chen Geng$^{3,*}$\\
Heng Zhang$^{1}$ & Jinhyung Ahn$^{4}$ & Caoliwen Wang$^{1}$ & Siyuan Chen$^{1}$\\
\end{tabular}\\
\begin{tabular}{c@{\hspace{1.6em}}c@{\hspace{1.6em}}c}
Donglai Xiang$^{5}$ & Jiajun Wu$^{3}$ & Peter Yichen Chen$^{1}$\\
\end{tabular}\\[0.8em]
$^{1}$University of British Columbia\qquad
$^{2}$University of Toronto\qquad
$^{3}$Stanford University\\
$^{4}$Yonsei University\qquad
$^{5}$NVIDIA
}

\begin{document}
\maketitle
{\renewcommand{\thefootnote}{$*$}\footnotetext{Corresponding authors.}}
\begin{strip}
\centering
\includegraphics[width=\linewidth]{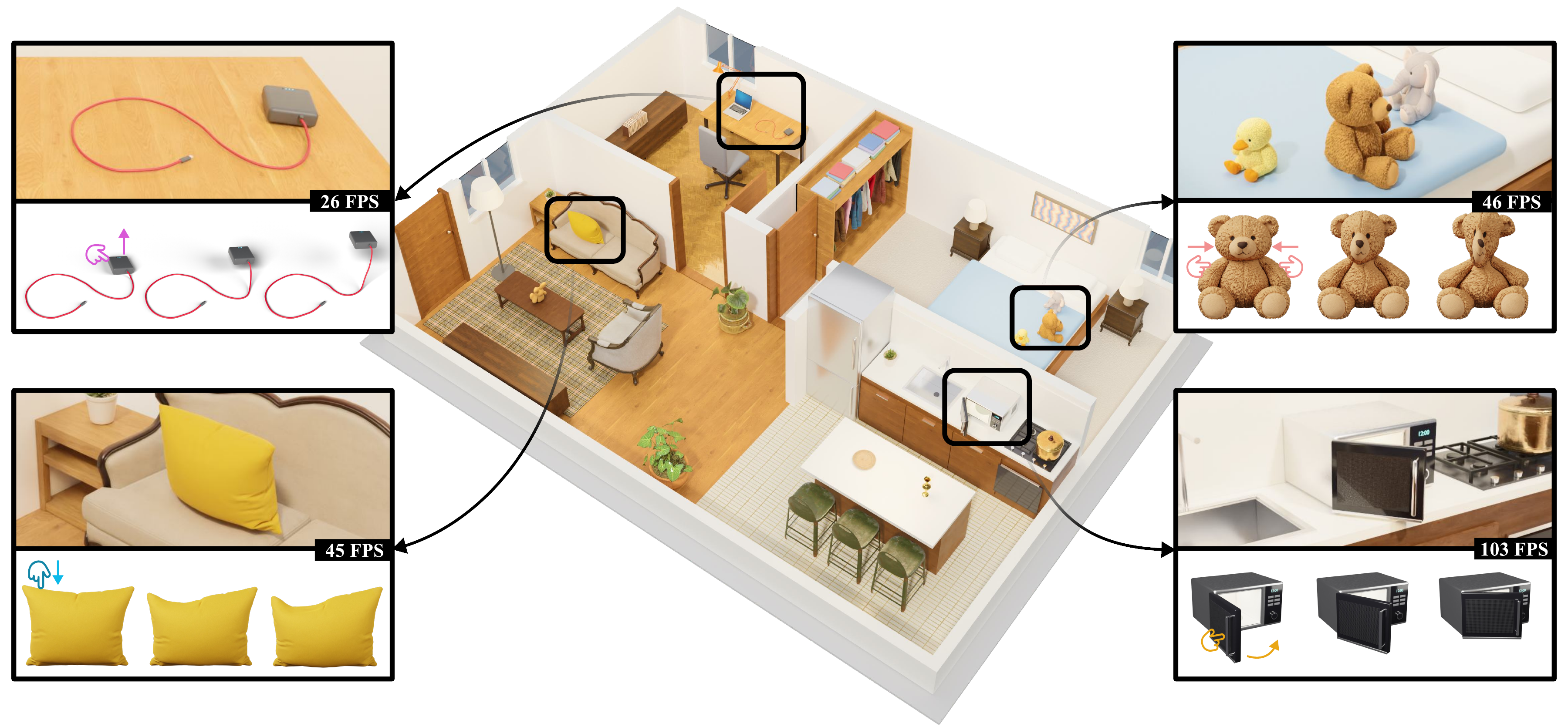}
\captionsetup{hypcap=false}
\captionof{figure}{\textbf{From a static 3D world to an interactive 4D world.} PRIO learns configuration priors from 4D trajectories that generalize to unseen objects without per-object training. Given only an object's static geometry, PRIO predicts a compact, object-specific configuration space that is linear, allowing physical responses to be generated efficiently using Lagrangian dynamics. By reusing precomputed matrices, PRIO enables objects throughout the scene to move and respond physically to user actions without further network inference, with nearly every object running at \textbf{over 30 FPS}.}
\captionsetup{hypcap=true}
\label{fig:teaser}%
\end{strip}
\input{sec/0_abstract}
\input{sec/1_intro}
\input{sec/2_related}
\input{sec/3_method}
\input{sec/4_interaction}

\input{sec/6_experiments}

\input{sec/7_conclusion}
{
    \small
    \bibliographystyle{ieeenat_fullname}
    \bibliography{main}
}
\end{document}

%% file: preamble.tex
\newcommand{\towa}[1]{{\color{magenta}[Towa: #1]}}
\newcommand{\vx}{\mathbf{x}}
\newcommand{\vX}{\mathbf{X}}
\newcommand{\vu}{\mathbf{u}}
\newcommand{\vf}{\mathbf{f}}

\newcommand{\mPhi}{\boldsymbol{\Phi}}

\makeatletter
\renewcommand\paragraph{\@startsection{paragraph}{4}{\z@}{0.5em plus 0.2em minus 0.1em}{-0.5em}{\normalfont\normalsize\bfseries}}
\makeatother

\usepackage{cuted} %
\usepackage{currfile} %
\usepackage{caption} %
\usepackage{titletoc} %

%% file: sec/0_abstract.tex
\begin{abstract}
3D worlds can now be captured, reconstructed, and generated at scale, yet the objects in them stay still: a pressed pillow does not compress, and a pulled door does not swing. Bringing these objects into \textbf{interactive} 4D world models requires a representation of how they can move, deform, and respond physically to user actions in real time. We present \textbf{PRIO}, a \textbf{p}re-trained \textbf{r}epresentation for \textbf{i}nteractive \textbf{o}bjects. Given static geometry, PRIO predicts an object-specific configuration space in a single forward pass. This space provides generalized coordinates for Lagrangian mechanics, allowing physical laws and user actions to govern motion. Configurations are linear in these coordinates, allowing the quantities needed by the dynamics to be precomputed. Dynamics then run at interactive rates without further network inference. Pre-trained on diverse 4D trajectories, PRIO learns priors over object configurations that generalize to unseen shapes. It supports meshes, point clouds, and 3D Gaussians, bringing objects in reconstructed and generated scenes into interaction. We demonstrate its use in real-to-sim workflows and robot manipulation. In addition, compared with existing 4D generation methods, PRIO generates dynamics faster and follows user actions more faithfully.
\end{abstract}

%% file: sec/1_intro.tex
\section{Introduction}
\label{sec:intro}

World models aim to predict how the world changes under interaction~\cite{worldmodels,genie,diamond}. Video-based world models now generate convincing futures from images and actions, but their output is pixels rather than objects, and every step requires an expensive generative model. Meanwhile, 3D worlds can be captured, reconstructed, and generated at scale~\cite{3dgs,lrm,trellis,hunyuan3d2,marble,gae}, and once created, their geometry and appearance stay consistent. Yet the objects in these worlds stay still. For people and robots to interact with them in a closed loop, acting on objects, observing the resulting motion, and adjusting their next actions, object dynamics must extend to any object in these worlds and respond physically in real time.

Existing approaches deliver only one of these two properties. Image- and text-conditioned methods generate dynamic 3D sequences from video priors or 4D data~\cite{dreamgaussian4d,animateanymesh} without taking physical actions as input, and physics-grounded methods let applied actions drive the generated motion through estimated or specified physical properties~\cite{physdreamer,pixie,physgm,physctrl,wonderplay,realwonder}, but producing each motion takes from a minute to hours of computation. Compact configuration spaces instead describe an object's motion through a few coordinates. Constructed per object from simulated trajectories or an elastic energy~\cite{licrom,simplicits}, they are linear, so the quantities needed by the dynamics are precomputed and interaction runs in real time, but nothing transfers to the next object. NeuROK~\cite{neurok} learns these spaces from large-scale 4D data and carries them to unseen objects, yet its spaces are nonlinear, so a large network must run at every moment of the motion. Pre-trained spaces transfer and linear spaces run in real time; an interactive 4D world needs both.

We present \textbf{PRIO}, a \textbf{p}re-trained \textbf{r}epresentation for \textbf{i}nteractive \textbf{o}bjects. Our key idea is to keep the configuration space linear, but to predict it for each object instead of constructing it. Given an object's static geometry, whether a mesh, a point cloud, or 3D Gaussians, PRIO predicts its \emph{configuration space}, i.e., the deformed shapes the object can take, each specified by a small set of \emph{coordinates}; this space is PRIO's representation of the object. Pre-trained on diverse 4D trajectories, PRIO learns priors over how objects move and deform, and predicts these spaces for unseen shapes without per-object training. Because the configurations are linear in the coordinates, the network runs only once per object; physical laws and user actions then drive the coordinates through Lagrangian dynamics~\cite{landau1960}, and every step reuses quantities computed at the start. Linearity also lets 4D data supervise the predicted space directly during pre-training, without a learned encoder. As a result, objects from capture, reconstruction, and generation pipelines respond to users and robots in real time (\cref{fig:teaser}).

In summary, we present \textbf{PRIO}, a model pre-trained on 4D trajectories that represents an object by a configuration space predicted from its static geometry, combining priors that generalize to unseen shapes with configurations that are linear. We demonstrate the practical utility of PRIO's learned linear configuration spaces by enabling interactive, physics-based manipulation of unseen objects. PRIO supports meshes, point clouds, and 3D Gaussians without repeated network inference, opening up applications to real-world objects and robot manipulation.

%% file: sec/2_related.tex
\section{Related Work}
\label{sec:related}

\paragraph{Interactive world models.}
World models learn how an environment evolves and how actions affect its future observations~\cite{worldmodels}. Video-based approaches generate interactive environments~\cite{genie,diamond,gamengen}, driving scenes~\cite{gaia1}, and world simulations~\cite{unisim,sora,cosmos}, and benchmarks examine their consistency with physical laws~\cite{kang2024howfar,physicsiq}. In an interactive 4D world, objects must instead change their 3D configurations consistently under applied actions; PRIO contributes an object-level representation for this in real time.

\begin{figure*}[t]
\centering
\includegraphics[width=1.0\textwidth]{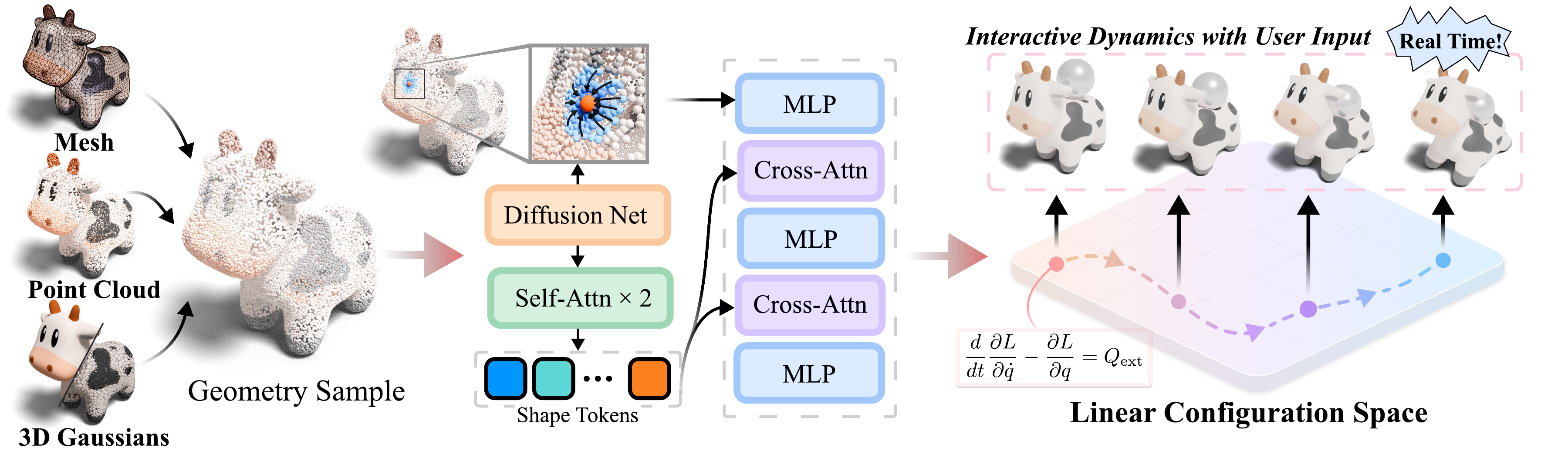}
\caption{\textbf{Overview of PRIO.} Given a mesh, point cloud or 3D Gaussians, PRIO encodes the geometry into shape tokens and predicts, at any point $X$, the fields $\Phi_\mathcal{G}(X)$ that span a linear configuration space. The network runs once per object; at inference, user actions drive the dynamics through the Euler--Lagrange equations in real time.}
\label{fig:overview}
\end{figure*}

\paragraph{Generating 4D dynamics.}
Dynamic 3D generation makes an object's evolution explicit in geometry. DreamGaussian4D~\cite{dreamgaussian4d} uses video-guided Gaussian deformation, while L4GM~\cite{l4gm} reconstructs dynamic Gaussian representations from video. AnimateAnyMesh~\cite{animateanymesh} learns mesh trajectories and generates animations conditioned on an initial shape and text. Physics-grounded approaches further connect dynamic generation with applied inputs. PhysGen~\cite{physgen} couples rigid-body simulation with image-to-video generation, PhysCtrl~\cite{physctrl} uses physics-grounded point trajectories to guide video synthesis, and WonderPlay~\cite{wonderplay} combines physical simulation with video generation for action-conditioned 3D scenes. Beyond generating dynamic sequences, learned simulators predict successive physical states using graph networks~\cite{gns,meshgraphnets}, particle-grid models~\cite{pgnd}, or particle transformers~\cite{worldparticle}. PRIO learns from 4D trajectories the configurations an object can take, while Lagrangian mechanics governs their evolution under physical conditions and applied actions.

\paragraph{Bringing visual objects into interaction.}
Connecting captured geometry with physical behavior makes existing 3D content interactive. Neural-field and Gaussian-based methods couple renderable geometry with continuum mechanics or spring-mass models and identify its physical properties~\cite{pacnerf,neuphysics,pienerf,physgaussian,omniphysgs,phys4dgen,gic,springgaus}, distill material fields from video generators~\cite{physdreamer}, or reconstruct deformable objects and garments from videos and images~\cite{phystwin,deformmaster,vid2sim,pgc,dress1to3,physavatar}. Feed-forward and vision-language methods predict physical properties or generate physically annotated assets~\cite{pixie,physgm,vomp,gaussianproperty,physsplat,physx3d}, and real-to-sim and agentic pipelines assemble simulation-ready worlds~\cite{splatsim,holodeck,physgen3d}. PRIO complements these efforts with a reusable configuration space for the resulting objects, predicted from their static geometry.

\paragraph{Compact configuration representations.}
Compact configuration representations describe object motion and deformation using a small number of coordinates. Parametric models provide such representations for categories such as faces and human bodies~\cite{morphable,smpl,smplx,npms,spams,nphm}, and learned pose-control and rigging methods for articulated objects~\cite{canor,unirig,puppeteer,skintokens}. Continuous deformation representations~\cite{windlifter} extend compact parameterizations to deformable objects. Such representations can be constructed for an individual object from simulated trajectories~\cite{crom,licrom,latentdyn,hoda,odddc} or a physical energy~\cite{simplicits,datafree,neuralmodes}. Among them, linear configuration maps~\cite{licrom,simplicits} admit precomputation and real-time dynamics, but each is constructed for one object and must be rebuilt for the next. PRIO keeps the configuration map linear and, by acquiring a shared configuration prior through pre-training on diverse objects, predicts it for each object without a skeleton or per-object training.

\paragraph{Learning representations across shapes.}
Several methods predict object representations from geometry. Shape Space Spectra~\cite{sss} learns deformation representations over a parametric shape family. NEO~\cite{neo} infers Laplacian eigenspaces, while NeuralSound~\cite{neuralsound} predicts vibration modes. Neural Jacobian Fields~\cite{njf} learns deformation fields, and PhysSkin~\cite{physskin} predicts skinning fields by self-supervision. Other methods learn to deform shapes across a collection through templates, cages, keypoints, or generative models~\cite{threedn,neuralcages,keypointdeformer,explorable}. NeuROK~\cite{neurok} learns a shape-conditioned nonlinear configuration space from 4D data, requiring network evaluations during dynamics; PRIO also learns from 4D trajectories, but its configuration map is linear, so no network runs during the dynamics.

%% file: sec/3_method.tex
\section{Pre-trained Linear Configuration Spaces}
\label{sec:method}

\subsection{Formulation}
\label{sec:formulation}

We describe an object by $n$ points with rest positions $X_1, \ldots, X_n \in \mathbb{R}^3$, which may be the vertices of a mesh, the points of a scan or the centers of 3D Gaussians; we call this set the geometry $\mathcal{G}$ of the object. We take the given static geometry as the object's rest state. When the object is pushed, dragged or grasped, its points move, and the result is a 4D sequence of \emph{configurations} $\vx^1, \vx^2, \ldots$, where $\vx^t \in \mathbb{R}^{3n}$ collects the positions of all points at time $t$; bold symbols denote such quantities stacked over the $n$ points.

Generating these trajectories requires describing which configurations an object can take and how to evolve among them. We model the object's configurations through a low-dimensional \emph{configuration space} $\mathcal{M}_\mathcal{G} \subset \mathbb{R}^{3n}$~\cite{crom}. Following LiCROM~\cite{licrom}, we approximate it with a continuous linear map; unlike LiCROM, which fits this map to each object, PRIO predicts its fields $\Phi_\mathcal{G} : \mathbb{R}^3 \to \mathbb{R}^{3 \times K}$ from the static geometry $\mathcal{G}$ and parameterizes configurations through the \emph{configuration map}
\begin{equation}
x(X) = X + \Phi_\mathcal{G}(X)\, q, \qquad q \in \mathbb{R}^K ,
\label{eq:statespace}
\end{equation}
where $q$ contains the object's \emph{latent coordinates} and $K$ is fixed by the network and shared by all objects ($K = 512$ in our experiments). Evaluating \cref{eq:statespace} at the $n$ points and stacking gives $\vx = \vX + \mPhi q$, where $\vX \in \mathbb{R}^{3n}$ stacks the rest positions and $\mPhi \in \mathbb{R}^{3n \times K}$ the predicted fields. The deformation $\vx - \vX$ therefore lies in the span of the columns of $\mPhi$, a linear subspace that approximates $\mathcal{M}_\mathcal{G}$.

Learning and interaction both operate in this subspace. During learning, observed configurations supervise the predicted fields through a least-squares fit of their latent coordinates (\cref{sec:pretraining}). During interaction, physical laws and user inputs drive the coordinates over time, generating a 4D trajectory of configurations (\cref{sec:interaction}). Because $\Phi_\mathcal{G}$ is continuous in $X$, these configurations can be evaluated at any points needed for interaction. Articulated objects use the same linear formulation, with $\Phi_\mathcal{G}$ parameterized through local affine transformations to capture joint motion; details are in the supplementary material. %

\subsection{Model Architecture}
\label{sec:architecture}

PRIO models $\Phi_\mathcal{G}$ conditioned on the geometry $\mathcal{G}$ with a single network (\cref{fig:overview}). It consists of a shape feature network, feature aggregation at query points, and a decoder that outputs $\Phi_\mathcal{G}(X)$.

\paragraph{Shape features.}
We sample $N_p$ points $P = \{p_j\}_{j=1}^{N_p}$ on the surface of the object, or take them from its point cloud or 3D Gaussian centers. An MLP maps the Fourier encoding~\cite{fourierfeatures} $\gamma(p_j)$ of each point to an initial feature, and $N_b$ DiffusionNet~\cite{diffusionnet} blocks refine these into point features $F = \{f_j\}_{j=1}^{N_p}$, $f_j \in \mathbb{R}^C$, by diffusing them over $P$ with a point-cloud Laplacian~\cite{sharp2020laplacian} at learned time scales, from local to global. We then select a subset using farthest point sampling~\cite{pointnet2}, iteratively choosing the point farthest from the selected set to cover the shape. Their features and position encodings pass through self-attention layers~\cite{transformer}, producing $N_t$ \emph{shape tokens} $T \in \mathbb{R}^{N_t \times C}$, in the spirit of point-set and particle tokenizers~\cite{shape2vecset,worldparticle}; each token summarizes one region of the object in the context of all others.

\paragraph{Query features.}
The shape features are defined only at the sampled points $P$, whereas we would like $\Phi_\mathcal{G}$ to be evaluated at arbitrary query points $X \in \mathbb{R}^3$. We therefore interpolate the features to each query point by a Gaussian-weighted average over its $k$ nearest sampled points $\mathcal{N}_k(X) \subset P$, in the spirit of feature interpolation in point networks~\cite{pointnet2}. This is done at $S$ scales, and the results are concatenated:
\begin{equation}
h_\mathcal{G}(X) = \Big[\, \textstyle\sum_{p_j \in \mathcal{N}_k(X)} w_s(X, p_j)\, f^{(s)}_j \,\Big]_{s=1}^{S},
\label{eq:read}
\end{equation}
where $f^{(s)}_j$ is the feature of $p_j$ smoothed by diffusion over $P$ at scale $s$, and $w_s(X, p_j)$ are Gaussian weights of learned width $\sigma_s(X)$, normalized over $\mathcal{N}_k(X)$ and tapered to zero at its boundary, so that points enter and leave the neighborhood with zero weight (details in the supplementary material). Because the Gaussian weights change continuously as $X$ moves, $h_\mathcal{G}(X)$ varies continuously with $X$, and so does $\Phi_\mathcal{G}$ decoded from it; this encourages spatially coherent deformation.

\paragraph{Decoder.}
A decoder $D$ maps the encoding of $X$ and its query feature, together with the shape tokens $T$, to the $3 \times K$ matrix
\begin{equation}
\Phi_\mathcal{G}(X) = D\big(\gamma(X),\, h_\mathcal{G}(X);\, T\big),
\label{eq:decoder}
\end{equation}
where MLP layers are interleaved with residual~\cite{resnet} cross-attention blocks~\cite{transformer,shape2vecset} that let the query attend to $T$, and the output head splits into $B$ branches that each produce $K/B$ columns. Query points never attend to one another, so the prediction at $X$ does not depend on which other points are queried: $\Phi_\mathcal{G}$ can be evaluated at any set of points and any resolution. Hyperparameter values and further details are given in the supplementary material.

\subsection{Learning from 4D Configurations}
\label{sec:pretraining}

PRIO is pre-trained on 4D configuration trajectories spanning different objects and interactions. Only the configurations are used, without additional physical labels or assumptions. For a frame $\vx^t$ of an object with rest state $\vX$, we write $\vu^t = \vx^t - \vX$ for its deformation, the motion of every point away from rest.

\paragraph{Optimal latent coordinates.}
Encoder--decoder methods~\cite{latentdyn,hoda,crom,neurok} rely on an encoder to find latent coordinates for an observation. The fit is therefore only as good as the coordinates the encoder finds, which may not give the best reconstruction the decoder can produce. In PRIO, the optimal coordinates can be \emph{directly solved for}. Because \cref{eq:statespace} is linear in $q$, finding them is a least-squares problem with a closed-form solution. Let $\mPhi$ stack the predicted fields at the observed points and $\vu$ their deformations. Then
\begin{equation}
q^\star = \arg\min_q \;\lVert \mPhi q - \vu \rVert^2 = \mPhi^{+} \vu ,
\label{eq:lsq}
\end{equation}
where $\mPhi^{+}$ is the pseudo-inverse. It returns a minimum-norm solution: for the given predicted fields, $q^\star$ guarantees an optimal reconstruction of the observed deformation. The fit thus measures how well the predicted space can explain the observation, without a learned encoder; a comparison with a learned encoder is given in the supplementary material. Gradients pass through this solution to train the space $\Phi_\mathcal{G}$ directly.

\paragraph{Training objective.}
Each training step takes an object $\mathcal{G}$ and one of its frames, observed at two sets of sample points, $\tilde{X}$ with $N_f$ points and $\tilde{X}'$ with $N_r$ points, which need not coincide. The fitted latent coordinates $q^\star$ are computed on $\tilde{X}$ by \cref{eq:lsq}, and the loss is evaluated on $\tilde{X}'$, so that $\Phi_\mathcal{G}$ has to reproduce the frame everywhere, not only at the fitted points. With $\mPhi'$ and $\vu'$ the stacked $\Phi_\mathcal{G}$ and the deformation at $\tilde{X}'$, the loss is
\begin{equation}
\mathcal{L} = \frac{1}{\bar{u}^2}\,\lVert \mPhi' q^\star - \vu' \rVert^2 + \lambda_\mathrm{d}\, \mathcal{L}_\mathrm{d} + \lambda_q \lVert q^\star \rVert^2 ,
\label{eq:loss}
\end{equation}
with weights $\lambda_\mathrm{d}$ and $\lambda_q$. The first term is the reconstruction error, normalized by the mean squared deformation $\bar{u}^2$ of the object. The second term keeps the predicted deformation smooth between sample points: with $J(X) = \partial\,(\Phi_\mathcal{G}(X)\, q^\star) / \partial X$ its spatial derivative and $\mathcal{E}(i)$ the nearest neighbors of sample point $X_i \in \tilde{X}'$,
\begin{equation}
\mathcal{L}_\mathrm{d} = \sum_i \sum_{j \in \mathcal{E}(i)} \frac{\lVert J(X_i)\,(X_j - X_i) - (u_j - u_i) \rVert^2}{\lVert X_j - X_i \rVert^2} .
\label{eq:deriv}
\end{equation}
The last term regularizes the magnitude of the fitted coordinates, similar to the latent-code prior of DeepSDF~\cite{park2019deepsdf}.

%% file: sec/4_interaction.tex
\section{Generating Interactive 4D Dynamics}
\label{sec:interaction}
PRIO turns a static object into one that responds to interaction: physical laws and user actions determine $q$, and its evolution generates the 4D motion.

\subsection{Lagrangian Dynamics}
\label{sec:lagrangian}
Through \cref{eq:statespace}, the coordinates $q$ determine the configuration of the entire object, so its motion is fully described by the trajectory $q(t)$. Treating $q$ as generalized coordinates, as in classical model reduction~\cite{pentland1989,witkin1990,barbic2005} and its learned counterparts~\cite{crom,licrom,neurok}, we obtain this trajectory from the Euler--Lagrange equations~\cite{landau1960}
\begin{equation}
\frac{d}{dt}\frac{\partial L}{\partial \dot q}
-\frac{\partial L}{\partial q}
=Q_\mathrm{ext},
\qquad
L(q,\dot q)=E_\mathrm{k}(\dot q)-V(q),
\label{eq:lagrange}
\end{equation}
where $V$ includes potentials such as elasticity and gravity. Because the configuration map is linear in $q$, point velocities are $\dot{\vx}=\mPhi\dot q$, and the kinetic energy is $E_\mathrm{k}=\tfrac12\dot q^\top M\dot q$, with $M=\mPhi^\top\mathbf{M}_x\mPhi$, where $\mathbf{M}_x$ is the mass matrix in physical space. For a nonlinear configuration map, the mass matrix generally depends on $q$, introducing velocity-dependent terms~\cite{neurok}. Here, $M$ is independent of $q$, so the equations reduce to
\begin{equation}
M\ddot q+\nabla_q V(q)=Q_\mathrm{ext}.
\label{eq:eom}
\end{equation}
Applied forces $\vf_\mathrm{ext}$ enter as $Q_\mathrm{ext}=\mPhi^\top \vf_\mathrm{ext}$. We integrate the dynamics under these inputs with implicit Euler; details are in the supplementary material.

\subsection{Fast Interaction through Linearity}
\label{sec:actions}
Because the positions depend linearly on the generalized coordinates, they have a fixed Jacobian and vanishing second derivatives,
\begin{equation}
\frac{\partial \vx}{\partial q}=\mPhi,
\qquad
\frac{\partial^2 \vx}{\partial q_i\,\partial q_j}=0.
\label{eq:linear_derivatives}
\end{equation}
We therefore precompute $\mPhi$ and, for a fixed mass distribution, $M$. The potential derivatives depend on the evolving configuration, and we project them onto the generalized coordinates with the same precomputed matrix at every step:
\begin{equation}
\nabla_q V=\mPhi^\top\nabla_{\vx} V,
\qquad
\nabla_q^2 V=\mPhi^\top(\nabla_{\vx}^2 V)\mPhi.
\end{equation}
These projections and the reconstruction of the 3D geometry can be batched on the GPU. Once PRIO predicts $\mPhi$, the time-stepping loop runs without further network inference.

%% file: sec/6_experiments.tex
\section{Experiments}
\label{sec:experiments}

\ifdefined\ARXIV
\begin{figure*}[t]
\centering
\includegraphics[width=0.9\textwidth]{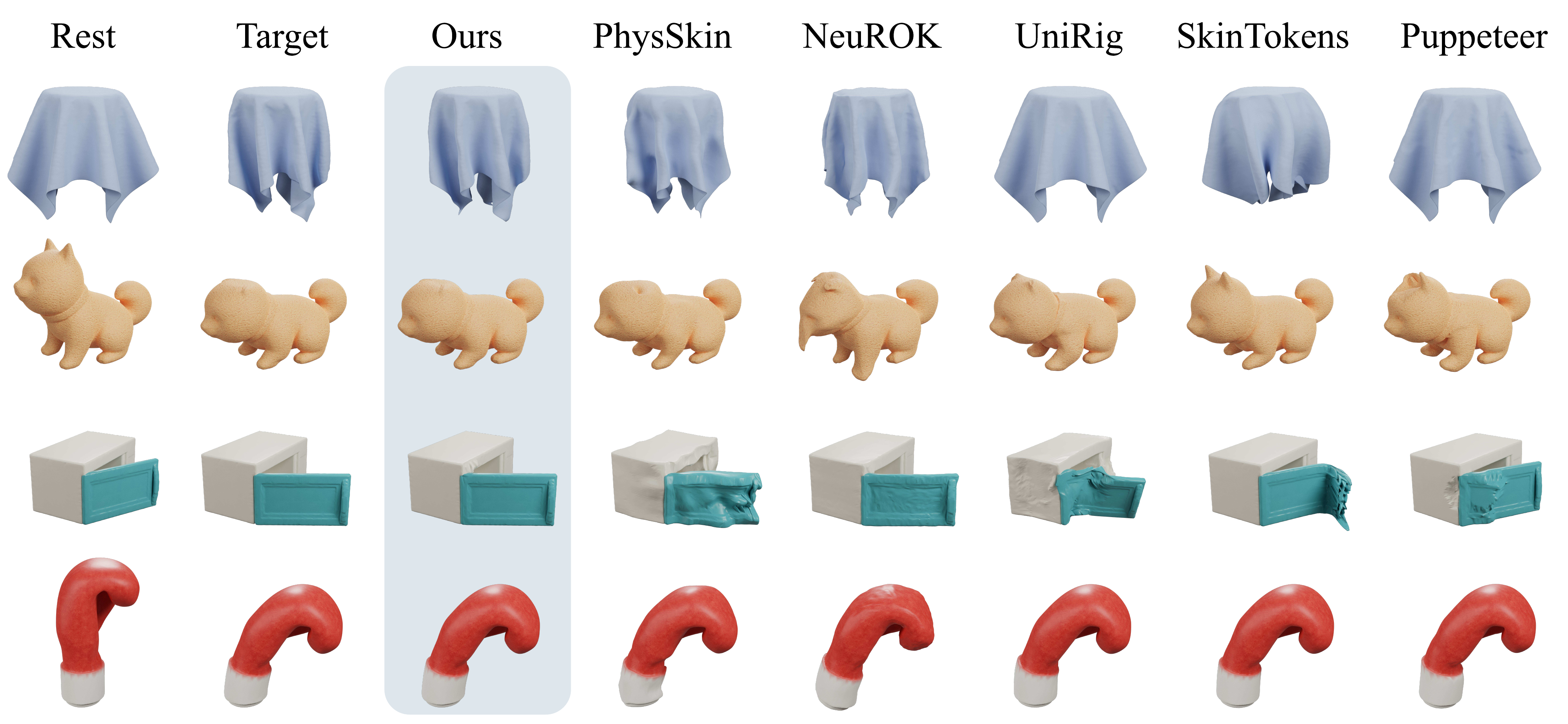}
\caption{\textbf{Qualitative inverse kinematics comparison.} We evaluate how well each method's predicted representation captures target configurations. Given an object's rest shape, we optimize its latent coordinates to fit each target and compare the resulting reconstructions. PRIO reproduces the targets most closely.}
\label{fig:inverse_kinematics}

\vspace{1mm}
\begin{minipage}[b]{\columnwidth}
\centering
\includegraphics[width=1\linewidth]{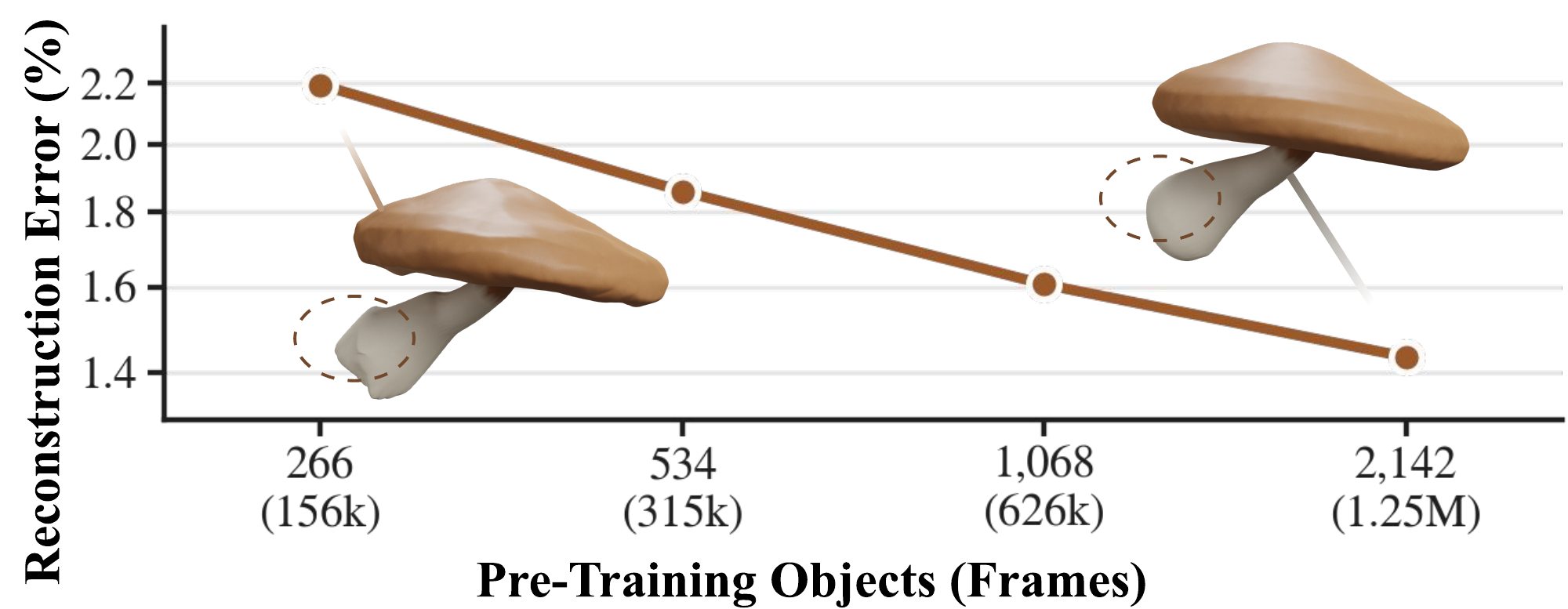}
\caption{\textbf{Effect of pre-training scale.} Reconstruction error on unseen shapes as the pre-training set grows from 266 to 2,142 objects under the same training budget. The error falls steadily, from 2.19\,\% to 1.43\,\%.}
\label{fig:pretrain}
\end{minipage}\hfill
\begin{minipage}[b]{\columnwidth}
\centering
\includegraphics[width=\linewidth]{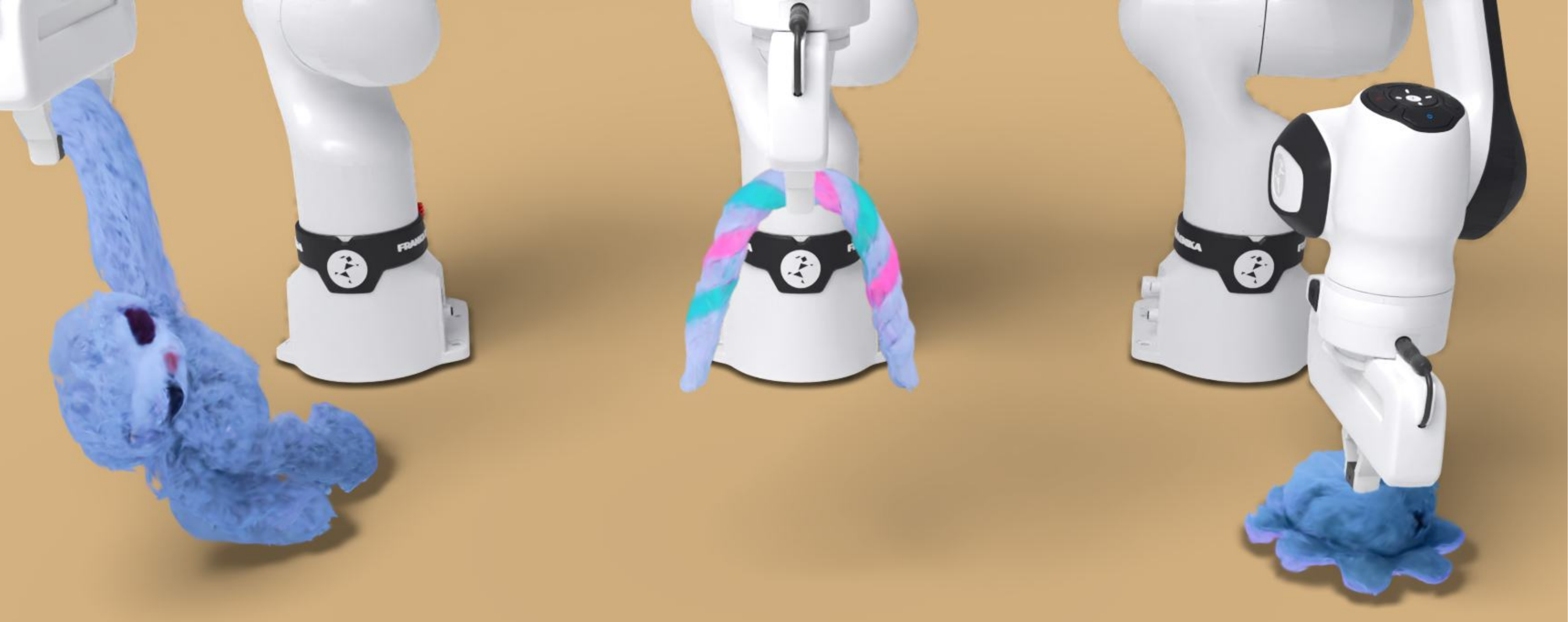}
\caption{\textbf{Interacting with real captured objects.}
PRIO enables physically driven responses to robot-arm manipulation for objects captured in Deform360~\cite{deform360}. Ropes bend and plush toys deform naturally without per-object training.}
\label{fig:deform360}
\end{minipage}
\end{figure*}
\else
\begin{figure*}[t]
\centering
\includegraphics[width=0.8\textwidth]{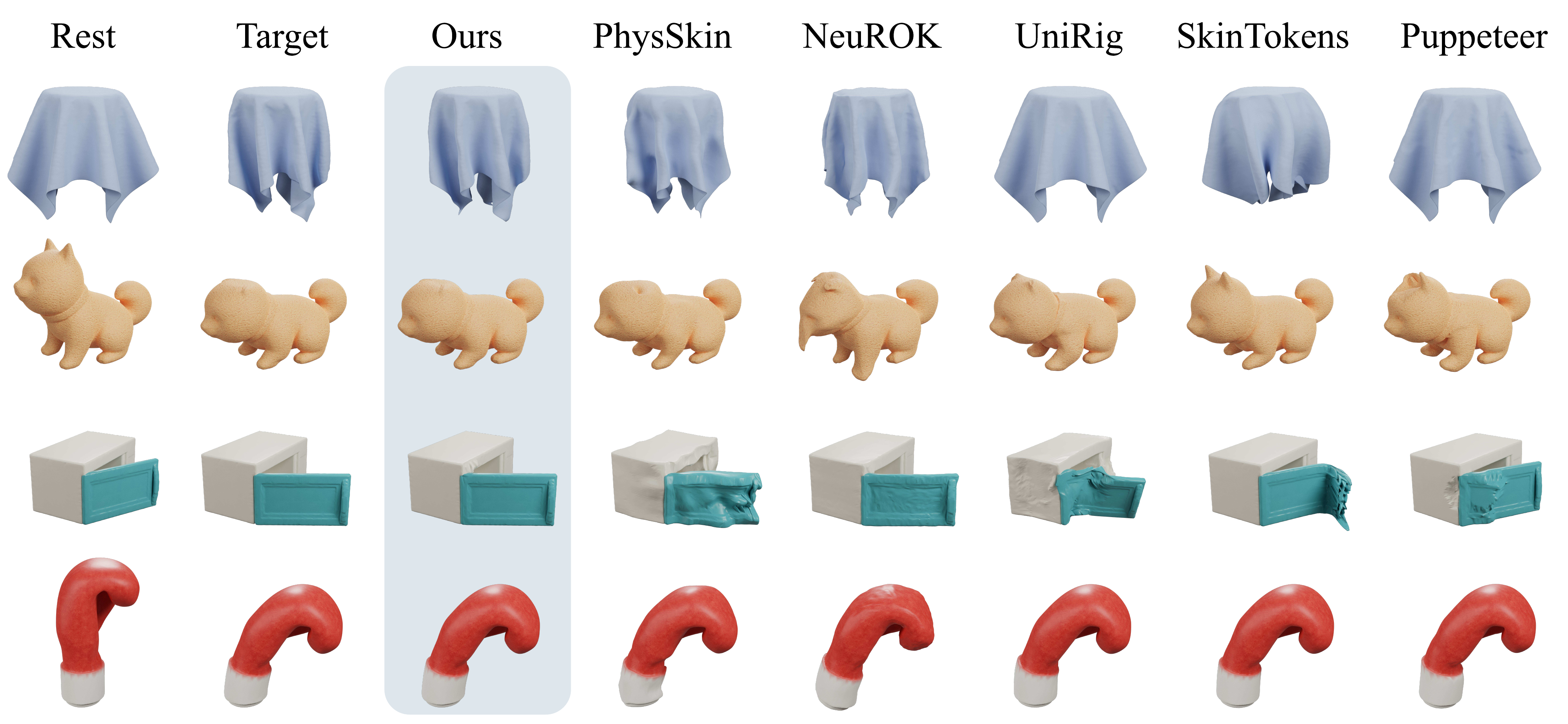}
\vspace{-2mm}
\caption{\textbf{Qualitative inverse kinematics comparison.} We evaluate how well each method's predicted representation captures target configurations. Given an object's rest shape, we optimize its latent coordinates to fit each target and compare the resulting reconstructions. PRIO reproduces the targets most closely.}
\label{fig:inverse_kinematics}

\vspace{3mm}
\begin{minipage}[t]{\columnwidth}
\centering
\includegraphics[width=1\linewidth]{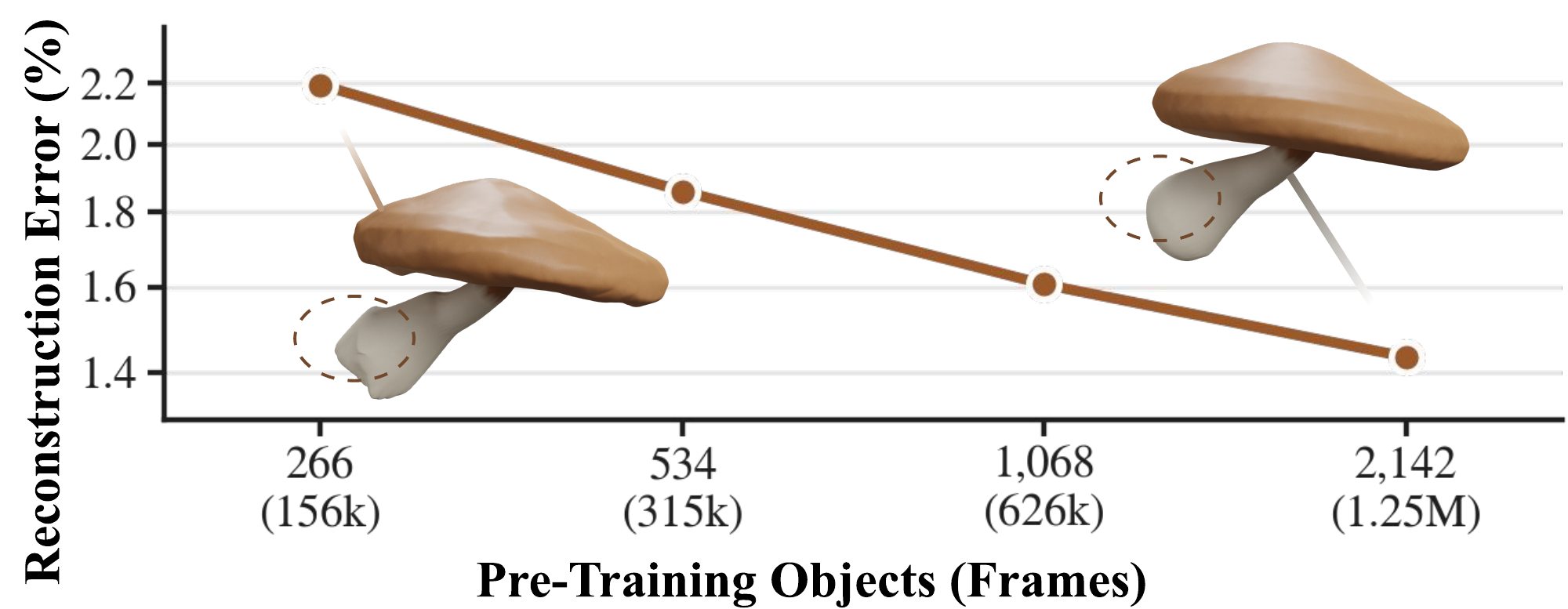}
\caption{\textbf{Effect of pre-training scale.} Reconstruction error on unseen shapes as the pre-training set grows from 266 to 2,142 objects under the same training budget. The error falls steadily, from 2.19\,\% to 1.43\,\%.}
\label{fig:pretrain}
\end{minipage}\hfill
\begin{minipage}[t]{\columnwidth}
\centering
\includegraphics[width=\linewidth]{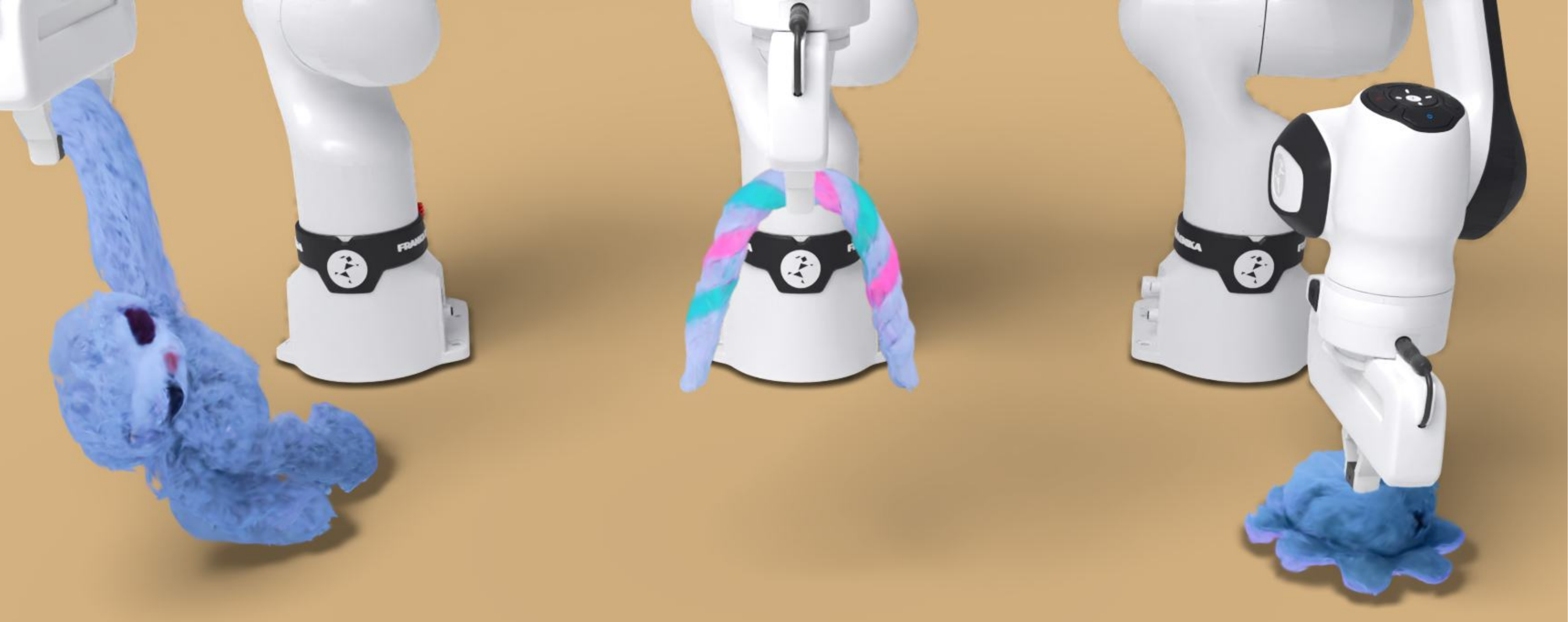}
\caption{\textbf{Interacting with real captured objects.}
PRIO enables physically driven responses to robot-arm manipulation for objects captured in Deform360~\cite{deform360}. Ropes bend and plush toys deform naturally without per-object training.}
\label{fig:deform360}
\end{minipage}
\end{figure*}
\fi

We show that PRIO combines generalizable configuration priors with real-time physical interaction. Pre-trained solely on simulated 4D trajectories and evaluated without per-object training, its predicted spaces capture the deformations of unseen shapes, improve with pre-training scale and transfer zero-shot to real captures (\cref{sec:exp_statespace,sec:exp_pretrain}), and enable real-time physical responses on meshes, point clouds and 3D Gaussians (\cref{sec:exp_input}). Against 4D generation methods, PRIO follows user actions more faithfully in a fraction of their time (\cref{sec:exp_4d_generation}). We also ablate key designs in \cref{sec:exp_ablation}; details of the pre-training data, additional results and comparisons are provided in the supplementary material. %
\subsection{Representation Quality}
\label{sec:exp_statespace}

A pre-trained configuration space is useful only if it can express how an unseen object deforms. Following CANOR~\cite{canor} and NeuROK~\cite{neurok}, we test this through inverse kinematics on unseen objects: each method builds an object's representation from its rest shape, and we fit deformed targets within it, measuring Chamfer $L_1$ and $L_2$ distances~\cite{psgn} and volumetric IoU. We compare with five learned object representations that span the common alternatives: skeletons with skinning weights, predicted by UniRig~\cite{unirig}, Puppeteer~\cite{puppeteer} and SkinTokens~\cite{skintokens}; neural skinning fields, predicted by PhysSkin~\cite{physskin}; and the nonlinear latent kinematic space that NeuROK~\cite{neurok} predicts with a conditional VAE.

\begin{table}[t]
\centering
\small
\setlength{\tabcolsep}{4pt}
\begin{tabular}{@{}lccc@{}}
\toprule
Method & Chamfer $L_1\downarrow$ & Chamfer $L_2\downarrow$ & IoU$\uparrow$ \\
\midrule
PRIO & \textbf{0.00522} & \textbf{0.000054} & \textbf{0.975} \\
PhysSkin~\cite{physskin} & 0.01299 & 0.000419 & 0.886 \\
SkinTokens~\cite{skintokens} & 0.02323 & 0.001732 & 0.829 \\
Puppeteer~\cite{puppeteer} & 0.03020 & 0.003153 & 0.791 \\
UniRig~\cite{unirig} & 0.03041 & 0.003001 & 0.784 \\
NeuROK~\cite{neurok} & 0.04300 & 0.004487 & 0.679 \\
\bottomrule
\end{tabular}
\caption{\textbf{Inverse kinematics comparison.} All methods are evaluated on unseen objects. PRIO fits the targets most closely on all three metrics.}
\label{tab:ik}
\end{table}

PRIO fits the targets most closely on all three metrics (\cref{tab:ik,fig:inverse_kinematics}): the space predicted from the rest shape alone captures the deformations of objects unseen in pre-training. Skinning-based representations blend the motion of a few handles, which can bend limbs but not express the local deformation of soft regions. NeuROK is trained largely on articulated objects, which explains its weaker fit on these deformable ones; we further compare with it on articulated objects, using the articulated extension of PRIO, in the supplementary material.

\subsection{Effect of Training Data}
\label{sec:exp_pretrain}

\begin{table}[t]
\centering
\small
\setlength{\tabcolsep}{5pt}
\ifdefined\ARXIV\begin{tabular*}{\linewidth}{@{\extracolsep{\fill}}lcccc@{}}\else\begin{tabular}{lcccc}\fi
\toprule
Pre-training scale & 266 & 534 & 1{,}068 & 2{,}142 \\
\midrule
Reconstruction error (\%) $\downarrow$ & 7.71 & 7.43 & 7.09 & \textbf{6.98} \\
\bottomrule
\ifdefined\ARXIV\end{tabular*}\else\end{tabular}\fi
\caption{\textbf{Zero-shot transfer to real captures.} Reconstruction error on 10 randomly sampled Deform360~\cite{deform360} objects for PRIO pre-trained on growing subsets of the pre-training set. Larger pre-training sets transfer better.}
\label{tab:deform360}
\end{table}

\begin{figure*}[t]
\centering
\ifdefined\ARXIV
\includegraphics[width=0.95\textwidth]{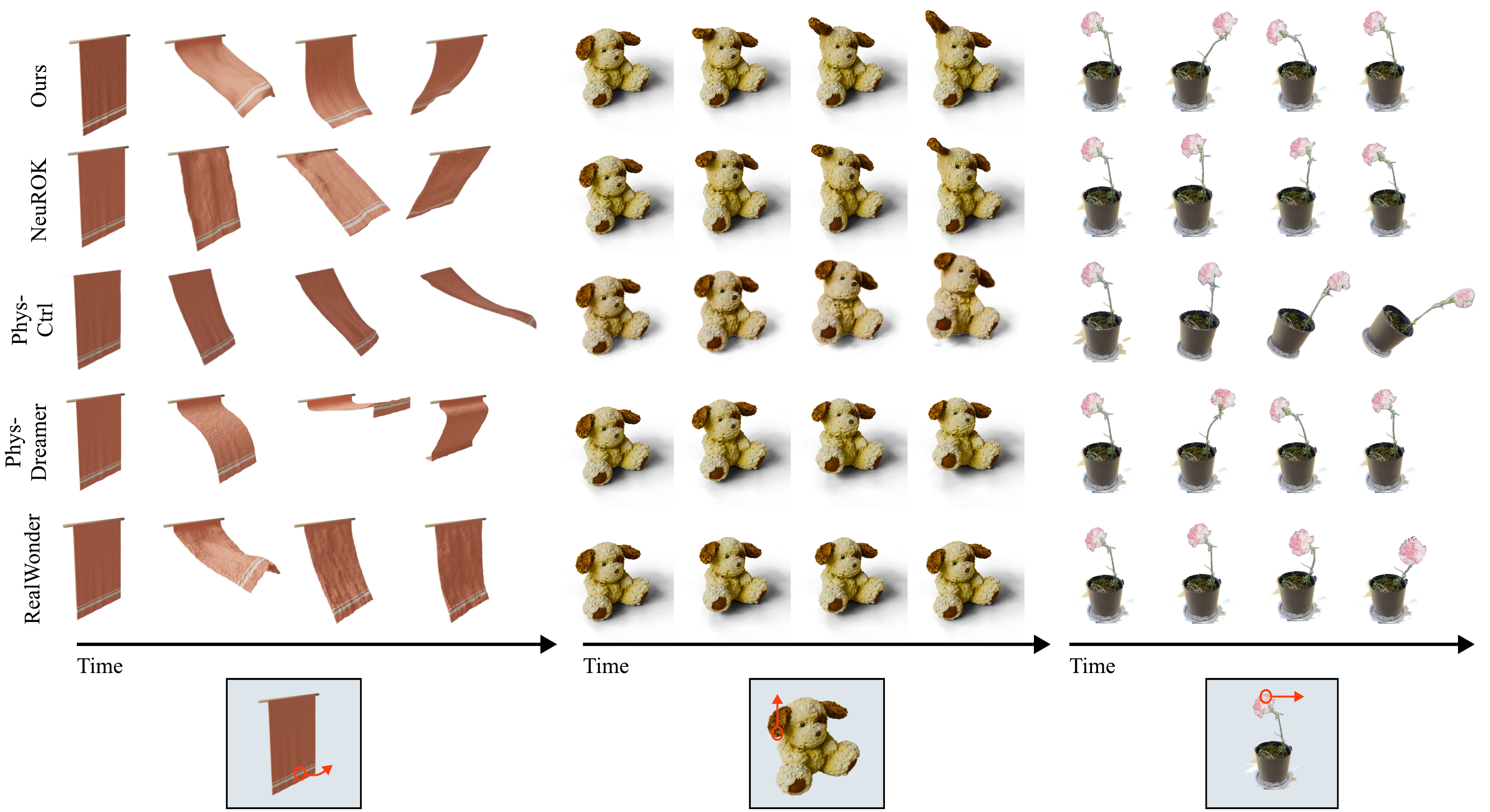}
\else
\includegraphics[width=0.95\textwidth]{fig/4d_baseline.png}
\vspace{-2mm}
\fi
\caption{\textbf{Qualitative comparison of interactive 4D dynamics.} We evaluate how naturally and plausibly objects move in response to user actions. PRIO deforms each object locally along the action and keeps it intact, whereas the baselines respond weakly or tear it.}
\label{fig:4d_generation}
\end{figure*}

\ifdefined\ARXIV
\begin{figure}[t]
\centering
\includegraphics[width=\linewidth]{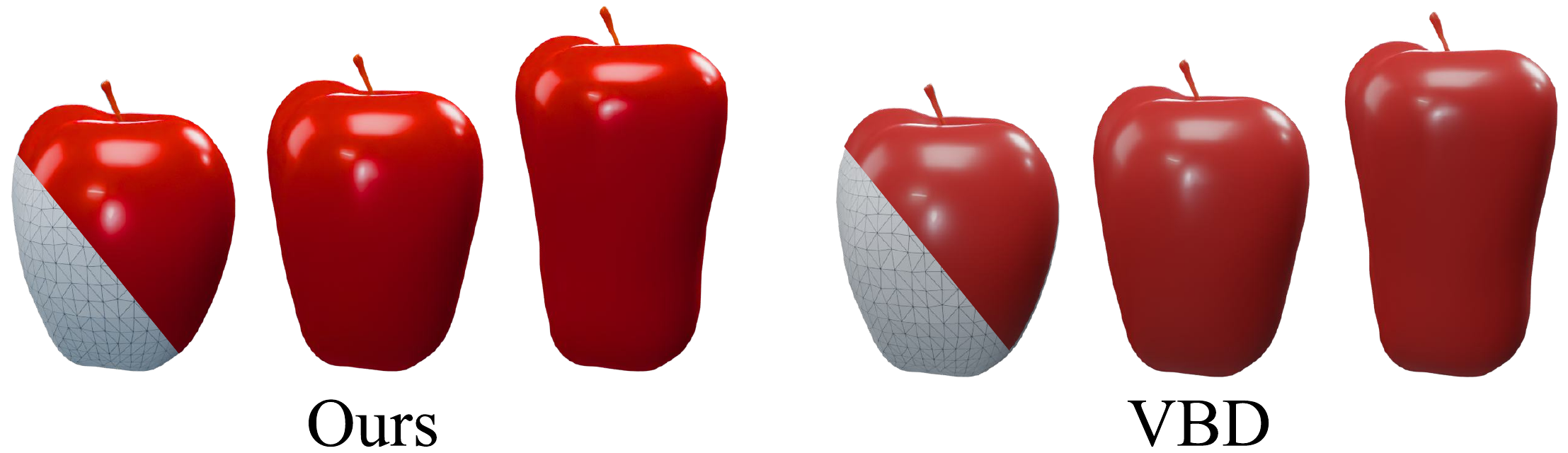}
\caption{\textbf{Speed and accuracy against VBD.} PRIO (left) and VBD~\cite{vbd} run to convergence (right) deform the same apple mesh on the same GPU. PRIO runs $7.1\times$ faster, while their surfaces differ by only 0.1\,\% of the object size on average.}
\label{fig:speed}
\end{figure}
\else
\begin{figure}[t]
\centering
\includegraphics[width=\linewidth]{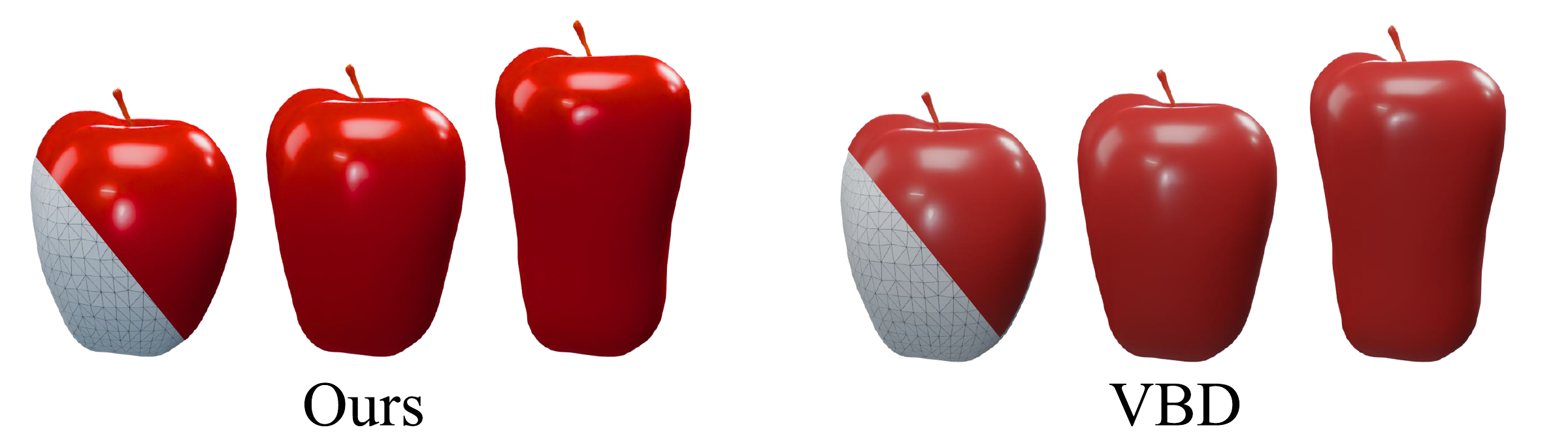}
\caption{\textbf{Speed and accuracy against VBD.} PRIO (left) and VBD~\cite{vbd} run to convergence (right) deform the same apple mesh on the same GPU. PRIO runs $7.1\times$ faster, while their surfaces differ by only 0.1\,\% of the object size on average.}
\label{fig:speed}
\end{figure}
\fi

\ifdefined\ARXIV
\begin{figure*}[t]
\centering
\includegraphics[width=0.9\textwidth]{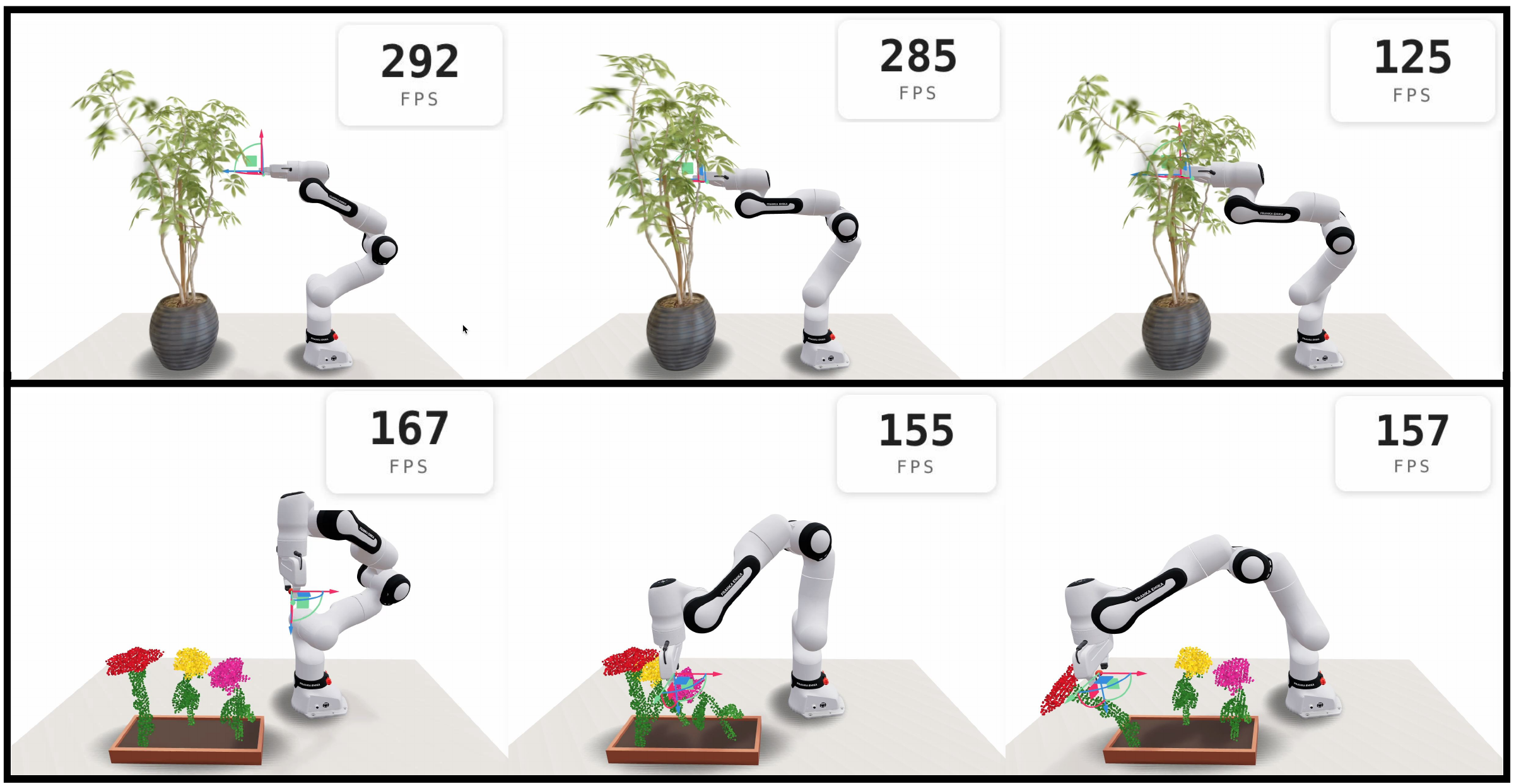}
\caption{\textbf{Live recordings of real-time interaction across 3D input formats.}
Screenshots of a user-controlled robot arm interacting with a 3D Gaussian ficus (top) and point-cloud flowers (bottom). The same pre-trained PRIO model simulates both objects' responses directly in their respective 3D representations, at 125--292\,FPS for the ficus and 155--167\,FPS for the flowers.}
\label{fig:gs_robot}
\end{figure*}
\else
\begin{figure}[t]
\centering
\includegraphics[width=\linewidth]{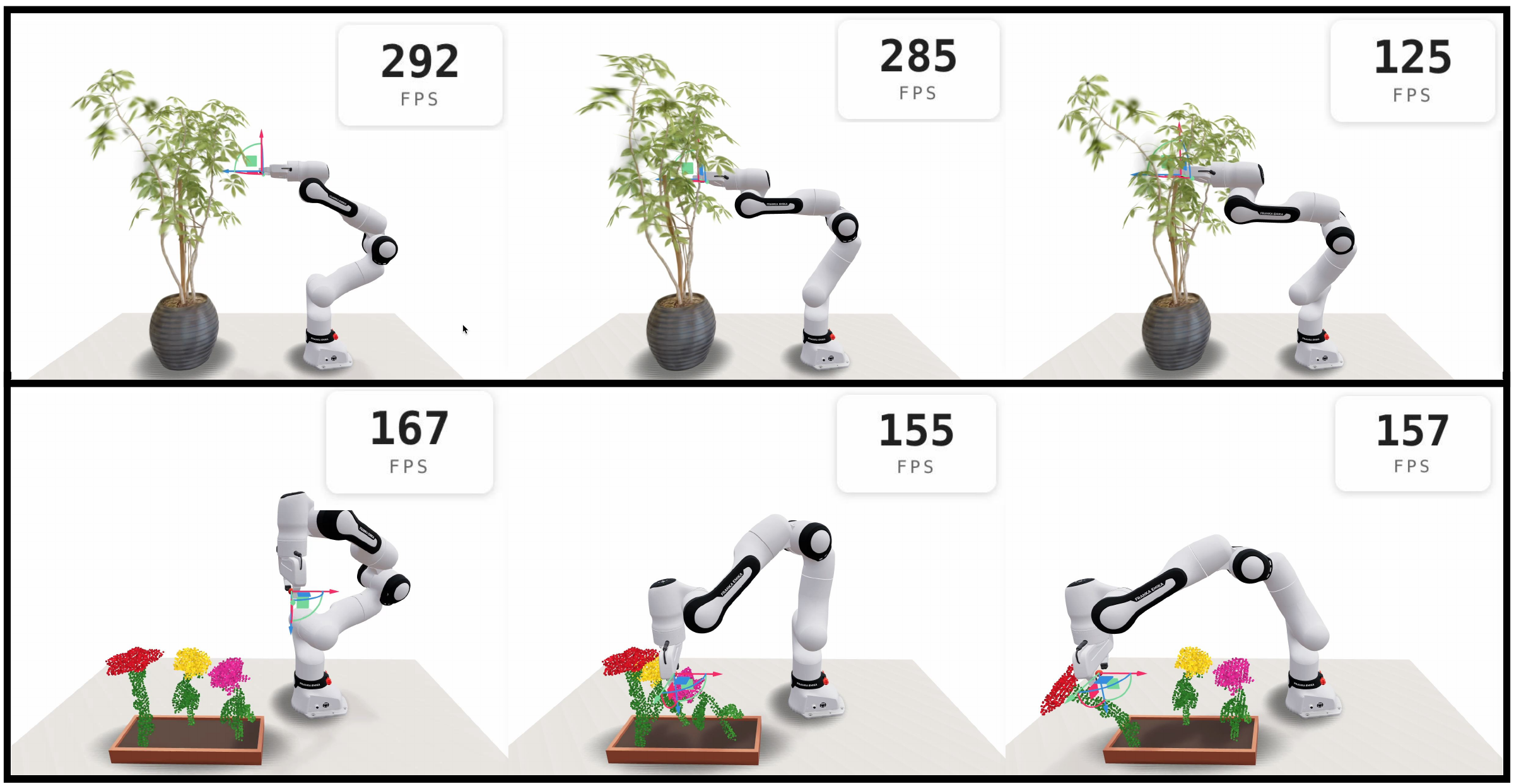}
\caption{\textbf{Live recordings of real-time interaction across 3D input formats.}
Screenshots of a user-controlled robot arm interacting with a 3D Gaussian ficus (top) and point-cloud flowers (bottom). The same pre-trained PRIO model simulates both objects' responses directly in their respective 3D representations, at 125--292\,FPS for the ficus and 155--167\,FPS for the flowers.}
\label{fig:gs_robot}
\end{figure}
\fi

A pre-trained space should improve as it sees more objects deform. To test this, we pre-train PRIO with the same budget of about 100k steps on nested subsets of the pre-training set, from 266 to 2,142 objects, and evaluate each model on the unseen shapes of \cref{sec:exp_statespace}, measuring the reconstruction error as the relative $L_2$ distance between the fitted and the observed deformation.

The error decreases steadily as the pre-training set grows (\cref{fig:pretrain}): more diverse 4D trajectories yield configuration spaces that better express how unseen objects deform.

The benefit of pre-training extends to real objects. On Deform360~\cite{deform360}, which captures real-world objects deforming under user interaction, PRIO recovers the captured deformations zero-shot through inverse kinematics, and the error again falls with pre-training scale (\cref{tab:deform360}). In \cref{fig:deform360}, a robot arm further manipulates these objects with PRIO: ropes bend and plush toys deform in response to its actions.

\subsection{Fast Interaction on Any 3D Input}
\label{sec:exp_input}
The same linearity that lets PRIO learn from 4D data also makes interaction fast. The matrices the dynamics require are cached once per object, and each step solves a dense system in a few hundred coordinates before mapping the motion back to the full geometry, without further network inference (\cref{sec:actions}). On the same apple mesh, simulation setting, and GPU, PRIO runs $7.1\times$ faster than VBD~\cite{vbd} run to convergence (10 iterations per substep), while producing nearly identical surface deformations (\cref{fig:speed}).

 Because the configuration map is spatially continuous, it can be evaluated at arbitrary points, so PRIO can also drive point clouds and 3D Gaussians. In \cref{fig:gs_robot}, a user steers a robot arm through two objects unseen in pre-training, a ficus represented by 3D Gaussians and flowers represented by point clouds; both bend, sway, and recover in real time as the arm passes through their foliage. PRIO thus delivers real-time physical responses regardless of how the object is represented.
\input{sec/6_4d_generation}

\subsection{Ablations}
\label{sec:exp_ablation}
We ablate the shape tokens (\cref{sec:architecture}) and $\mathcal{L}_\mathrm{d}$ (\cref{eq:deriv}) in \cref{tab:ablation} on the same unseen shapes, also reporting a sparse IK error, for which the coordinates are fitted to a sparse set of 256 random surface points and the error is measured over the whole surface. Both designs help the space fill in unobserved regions (details in the supplementary material).

\begin{table}[t]
\centering
\small
\ifdefined\ARXIV\begin{tabular*}{\linewidth}{@{\extracolsep{\fill}}lcc@{}}\else\begin{tabular}{@{}lcc@{}}\fi
\toprule
Variant & Reconstruction (\%) $\downarrow$ & Sparse IK (\%) $\downarrow$ \\
\midrule
PRIO & \textbf{1.43} & \textbf{6.61} \\
w/o shape tokens & 2.63 & 8.58 \\
w/o $\mathcal{L}_\mathrm{d}$ & 1.50 & 8.38 \\
\bottomrule
\ifdefined\ARXIV\end{tabular*}\else\end{tabular}\fi
\caption{\textbf{Ablation study.} Impact of shape tokens and $\mathcal{L}_\mathrm{d}$ on reconstruction and sparse IK accuracy. Both designs reduce the sparse IK error.}
\label{tab:ablation}
\end{table}

\begin{figure}[t]
\centering
\includegraphics[width=\linewidth]{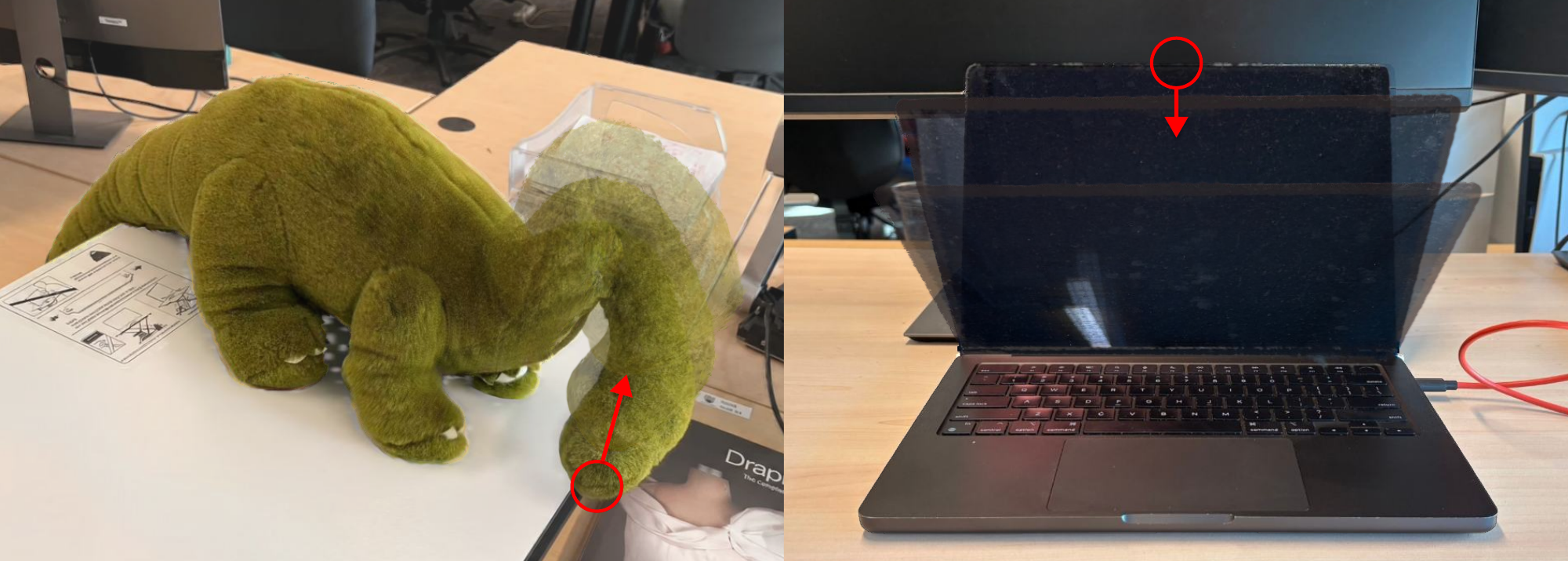}
\caption{\textbf{Interactive real-world objects.} Objects reconstructed from photographs deform under user actions; overlaid frames show the motion. Each responds at interactive rates without per-object training.}
\label{fig:cases}
\end{figure}

\begin{figure}[t]
\centering
\includegraphics[width=\linewidth]{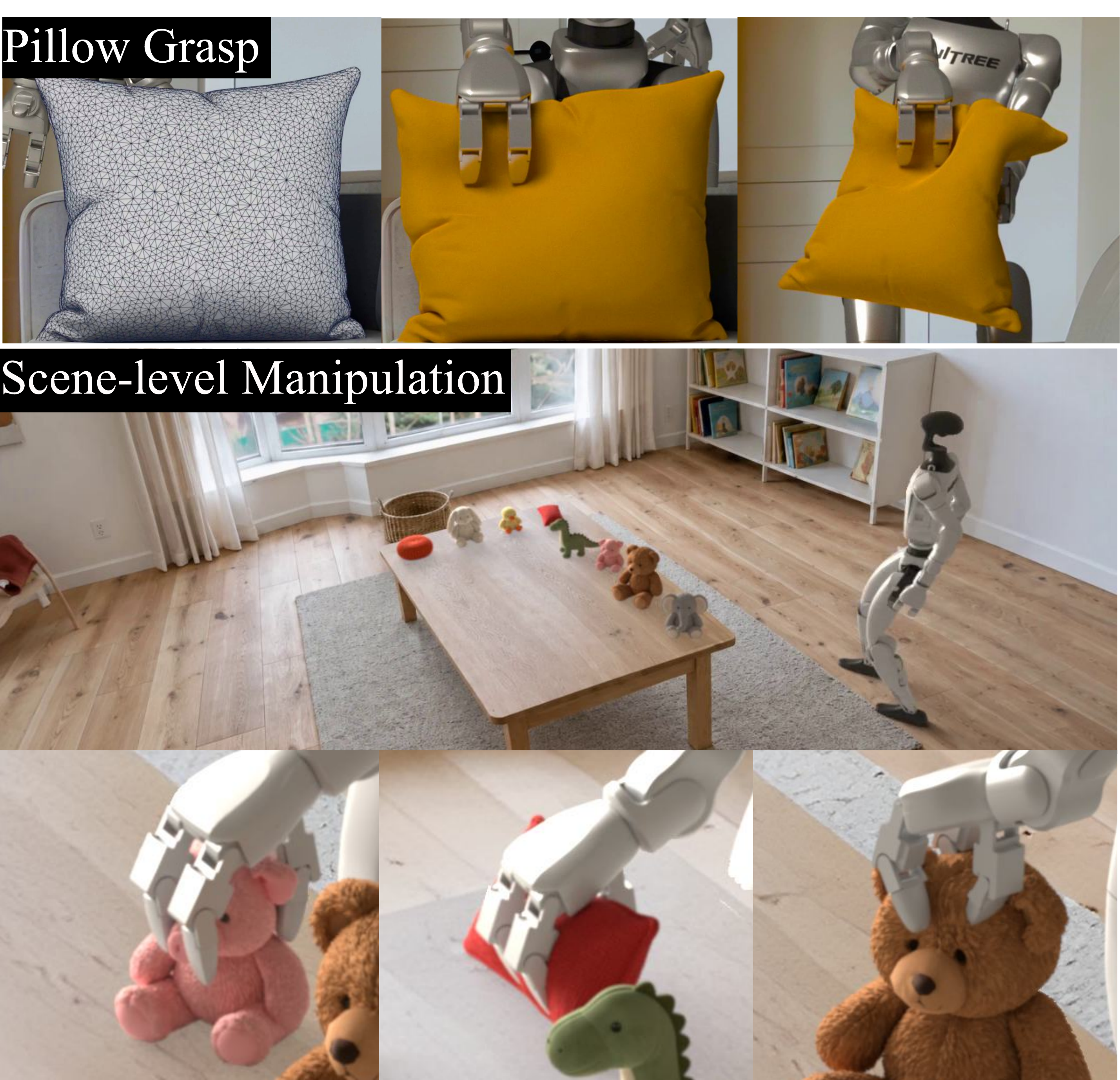}
\caption{\textbf{Robot interaction.} (a) Starting from the pillow's surface mesh, PRIO models the deformation under a robot grasp at 7.6\,ms per step. (b) A humanoid robot grasps eight plush toys in the same scene, all simulated together at 25\,ms per step.}
\label{fig:robot_pillow}

\end{figure}

\section{Applications}
\ifdefined\ARXIV
Both applications below place objects in a loop with live actions, which requires a configuration space without per-object training and dynamics fast enough to keep up.
\fi

\subsection{Interactive Real-World Objects}
\label{sec:exp_real}
A photograph of an object becomes an object that responds to touch (\cref{fig:cases}). PRIO predicts the configuration space of each reconstructed object without per-object training; the workflow is detailed in the supplementary material. None of these objects is in the pre-training set, yet each responds to user actions at interactive rates.

\subsection{Robot Interaction}
\label{sec:exp_robot}
Robot manipulation requires deformable objects that respond to contact at the robot's pace, often many at once. Given the robot's motion, PRIO simulates the pillow in \cref{fig:robot_pillow} (a), with a 145k-triangle surface, at 7.6\,ms per step on a single GPU, and the eight plush toys grasped by a humanoid robot in \cref{fig:robot_pillow} (b), nearly 1M triangles in total, together at 25\,ms per step on average. At this cost, the pre-trained prior still captures natural responses: folds form around the gripper on the pillow, and the plush toys deform where they are grasped.

%% file: sec/6_4d_generation.tex
\subsection{Interactive 4D Dynamics Generation}
\label{sec:exp_4d_generation}
\begin{table}[t]
\centering
\small
\setlength{\tabcolsep}{4pt}
\begin{tabular}{@{}lcccr@{}}
\toprule
Method & SA $\uparrow$ & PC $\uparrow$ & VQ $\uparrow$ & Time $\downarrow$ \\
\midrule
PhysDreamer~\cite{physdreamer} & 3.33 & 3.22 & 3.94 & 4.6\,h \\
RealWonder~\cite{realwonder}  & 3.39 & 3.17 & 3.44 & 2.8\,min \\
PhysCtrl~\cite{physctrl}      & 2.58 & 1.89 & 2.89 & 68\,s \\
NeuROK~\cite{neurok}          & 4.33 & 3.72 & 3.97 & 34\,min \\
\midrule
PRIO & \textbf{5.00} & \textbf{4.53} & \textbf{4.94} & \textbf{36\,s} \\
\bottomrule
\end{tabular}
\caption{\textbf{Quantitative comparison on 4D dynamics and generation time.} VLM scores (1--5) and end-to-end time to produce the 4D motion. PRIO leads in all metrics with the fastest runtime.}
\label{tab:4d}
\end{table}

Because PRIO generates dynamics by physics rather than by a generative model, it should follow user actions faithfully and at low cost. We give all methods the same three objects and actions: a cloth pulled at a corner, a plush toy lifted by its ear, and a flower pulled at its head. PhysDreamer~\cite{physdreamer} estimates physical properties from video priors, PhysCtrl~\cite{physctrl} generates force-conditioned point trajectories, RealWonder~\cite{realwonder} guides video generation with simulation previews, and NeuROK~\cite{neurok} evolves the motion in a nonlinear configuration space. A shape and an action do not determine a unique motion, so following PhysCtrl, a VLM scores semantic adherence (SA), physical commonsense (PC), and video quality (VQ) from 1 to 5. We also report the average end-to-end generation time. Details are in the supplementary material.

PRIO scores highest and generates fastest (\cref{tab:4d,fig:4d_generation}). PhysCtrl moves the whole object rigidly, and the other generative baselines respond weakly, tear or collapse. NeuROK follows the actions more closely but is much slower, since its network runs throughout the motion. PRIO follows the action and keeps each object intact.

%% file: sec/7_conclusion.tex
\section{Conclusion}
\label{sec:conclusion}

We have introduced PRIO, a pre-trained representation that brings static 3D objects into interactive 4D world models by unifying cross-object transfer and real-time physical interaction in a single configuration space. Predicted from static geometry without per-object training, the space captures the deformations of unseen shapes, improves with pre-training scale and transfers to real captures; because it is linear, the dynamics are precomputed once per object and user actions drive the object without further network inference, on meshes, point clouds and 3D Gaussians alike. Together, these properties make reconstructed and generated 3D content interactive. As a limitation, the predicted configurations are continuous and low-dimensional, so topology changes such as tearing and fracture, as well as behaviors absent from pre-training such as fluids, lie outside them. Broadening the pre-training data is a natural next step toward worlds that people and robots can act in.